\documentclass[10pt,twocolumn,letterpaper]{article}

\usepackage{iccv}
\usepackage{times}
\usepackage{epsfig}
\usepackage{graphicx}
\usepackage{amsmath}
\usepackage{amssymb}
\usepackage{algorithm}
\usepackage{algorithmic}
\usepackage{enumitem}
\usepackage{footnote}

\usepackage{soul}
\newcommand\given[1][]{\:#1\vert\:}
\newcommand{\searchmethod}{GRAM\xspace}

\usepackage{authblk}

\usepackage[pagebackref=true,breaklinks=true,letterpaper=true,colorlinks,bookmarks=false]{hyperref}

\iccvfinalcopy 

\def\iccvPaperID{3736} 

\ificcvfinal\fi
\begin{document}

\title{SwiftNet: Using Graph Propagation as Meta-knowledge to Search Highly Representative Neural Architectures}
\author[1]{Hsin-Pai Cheng}
\author[1]{Tunhou Zhang}
\author[1]{Yukun Yang}
\author[3]{Feng Yan}
\author[1]{Shiyu Li}
\author[2]{Harris Teague}
\author[1]{Hai Li }
\author[1]{Yiran Chen}
\affil[1]{ECE Department, Duke University, Durham, NC 27708}
\affil[2]{Qualcomm AI Research, 5775 Morehouse Drive, San Diego, CA 92121}
\affil[3]{CSE Department, University of Nevada, Reno, NV 89557}

\maketitle

\begin{abstract}
Designing neural architectures for edge devices is subject to constraints of accuracy, inference latency, and computational cost. Traditionally, researchers manually craft deep neural networks to meet the needs of mobile devices. 
Neural Architecture Search (NAS) was proposed to automate the neural architecture design without requiring extensive domain expertise and significant manual efforts.   
Recent works utilized NAS to design mobile models by taking into account hardware constraints  
and achieved state-of-the-art accuracy with fewer parameters and less computational cost measured in Multiply-accumulates (MACs). 
To find highly compact neural architectures, existing works relies on predefined cells and directly applying width multiplier,
which may potentially limit the model flexibility, reduce the useful feature map information, and cause accuracy drop. 
To conquer this issue, we propose GRAM (GRAph propagation as Meta-knowledge) that
adopts fine-grained (node-wise) search method and accumulates the knowledge learned in updates into a meta-graph. 
As a result, GRAM can enable more flexible search space and achieve higher search efficiency. 
Without the constraints of predefined cell or blocks, 
we propose a new structure-level pruning method to remove redundant operations in neural architectures.  
SwiftNet, which is a set of models discovered by GRAM, outperforms MobileNet-V2 by 2.15$\times$ higher accuracy density and 2.42$\times$ faster with similar accuracy. Compared with FBNet, SwiftNet reduces the search cost by 26$\times$ and achieves 2.35$\times$ higher accuracy density and 1.47$\times$ speedup while preserving similar accuracy. SwiftNet can obtain 63.28\% top-1 accuracy on ImageNet-1K with only 53M MACs and 2.07M parameters. 
The corresponding inference latency is only 19.09 ms on Google Pixel 1. 
\end{abstract}

\section{INTRODUCTION}

\begin{figure}[t]
\begin{center}
   \includegraphics[width=0.9\linewidth]{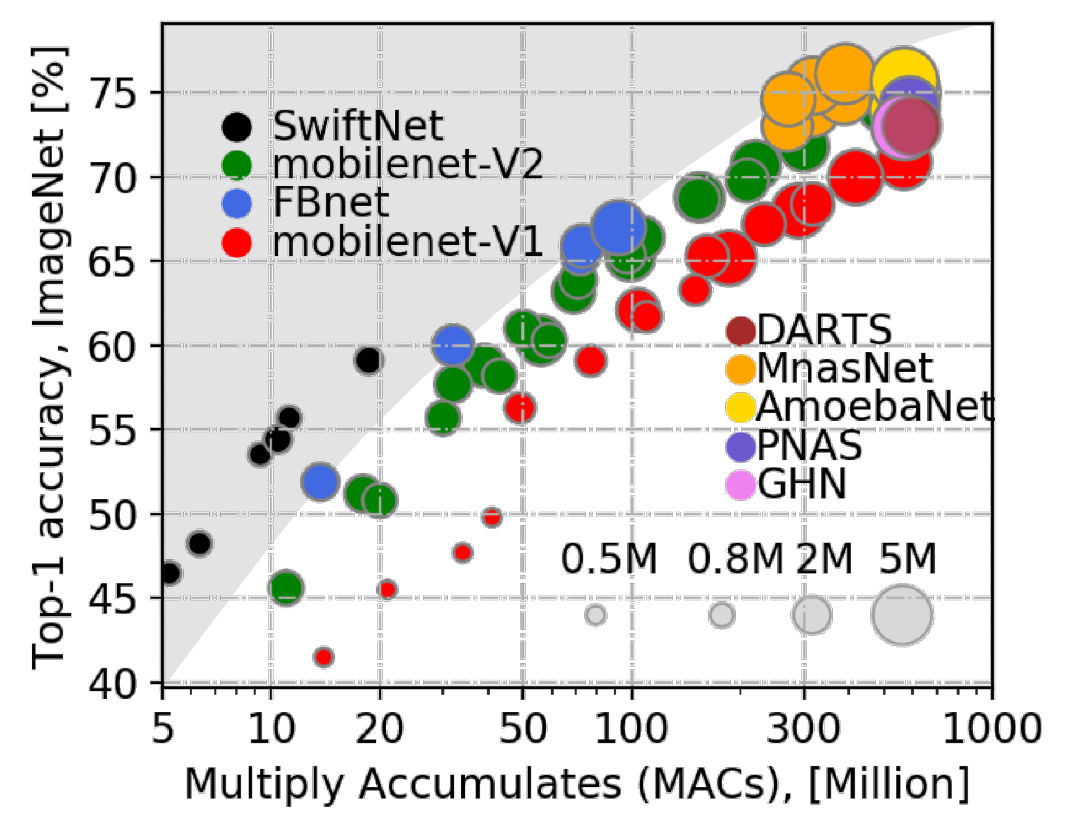}
   \end{center}
   \vspace{-1.5em}
   \caption{ImageNet-1K top-1 accuracy density vs Model MACs. SwiftNet achieves a better Pareto-optimal of accuracy-MAC curve of other state-of-the-art models with significant smaller size. Experiments are detailed in Section~\ref{sec:setup}.}
\label{fig:MAC_density}
\end{figure}


Deep Neural Networks (DNNs) are able to achieve state-of-the-art performance in many cognitive applications, including computer vision~\cite{he2017mask}, speech recognition~\cite{hinton2012deep}, and natural language processing~\cite{devlin2018bert, lample2018phrase}. 
The architectures of DNNs are continuously becoming deeper and wider with many new operations and activation function designs being constantly invented. 
Such dynamics make manual design of DNNs very challenging, especially when extensive domain expertise and experience are needed. 

Neural Architecture Search (NAS) was recently proposed to automate the search for neural network architecture~\cite{zoph2016neural}. 
Evolutionary algorithms~\cite{real2017large, elsken2018efficient}, reinforcement learning~\cite{zoph2018learning}, and differentiable algorithms~\cite{liu2018darts, wu2018fbnet} are the trending methods for performing NAS and obtaining networks; the performance of which outperforms the best manually-crafted DNN models~\cite{howard2017mobilenets}.
NAS was particularly studied for searching large-scale neural networks that consume a large amount of computing power and time~\cite{zoph2018learning}. 
However, in edge computing applications such as autonomous vehicles, disaster response, and health-care informatics, etc., many new challenges emerge, including limited computing power, short learning ramp-up time, real-time inference response, and quick adaptation to new tasks. 
Some recent studies have started to investigate these emerging challenges by taking into account hardware resource constraints, device runtime latency, and FLOPs during neural architecture design~\cite{gordon2018morphnet, cai2018proxylessnas, wu2018fbnet, dai2018chamnet}.
Although these techniques are able to reduce inference latency and/or computational cost, they often induce a significant accuracy drop when the model size is small, e.g., less than 20M MACs \cite{howard2017mobilenets, sandler2018mobilenetv2, wu2018fbnet}.


The accuracy drop is due to: (1) the predefined cell having less flexibility for small model design and (2) directly applying width multiplier largely reduces the useful feature map information.
 Figure~\ref{fig:MAC_density} compares the difference of the architectures we discovered with previous works. Previous works are related in that they are characterized by their similar Pareto distribution. Our discovered architectures go beyond the Pareto-optimal of previous works.

To overcome the above limitations of existing NAS approaches, we propose a new NAS methodology called \searchmethod (GRAph propagation as Meta-knowledge), which aims to achieve high performance, rapid and efficient searches, good adaptability, and multi-objective searches (e.g., resource-aware inference).
Our approach builds on abstracting the operation in computational Directed Acyclic Graph (DAG) as nodes and the flow of tensors into network layers as passing through an edge connecting two nodes. 
In this way, the problem of finding the optimal neural architecture in the full search space is equivalent to finding the optimal sub-DAG in the complete DAG.
There are two key components in our work: meta-graph and structure-level pruning.
Existing NAS approaches cannot retain the learned knowledge in previous updates when a new search starts, making the search process inefficient and with poor adaptability.
Thus we develop a new node-based search method that continuously accumulates the learned neural architecture knowledge into a meta-graph. 
To support resource constraints on edge devices, we propose a novel structure-level pruning method by removing edges with lower weight values in the complete DAG to produce resource-aware models.

Compared with traditional cell-based search methods, the proposed node-based method has much less constraint in search space and thus has significantly higher accuracy density. 
\searchmethod also enables a faster, more efficient search process as the learned information is accumulated in meta-graph instead of end-to-end learning as in existing NAS methods. With graph modularization, \searchmethod has more flexibility to accumulate and preserve knowledge from past training experiences and generalize for unseen architecture without any constraints.
In addition, \searchmethod has great adaptability, thanks to the meta-graph as it preserves the previously learned information which can be effectively reused for new tasks.
Finally, the proposed structure-level pruning is better than traditional cell-based pruning as the latter cannot identify structural redundancy while our method is capable of removing it in the trained meta-graph. It also facilitates changing objectives as it can be done by simply changing the meta-graph update rules. 
We conduct extensive experimental evaluations on a wide range of image classification applications, and the results show that SwiftNet achieves state-of-the-art accuracy density (8.90), which is 2.15$\times$ more than MobileNet-V2 (4.14) \cite{sandler2018mobilenetv2} and 2.35$\times$ more than FBNet (3.79) \cite{wu2018fbnet}. The latency of SwiftNet is 2.42$\times$  lower than MobileNet with similar accuracy. Our search cost is 26$\times$ lower than FBNet, and the latency of SwiftNet is 1.47$\times$ lower than FBNet with similar accuracy. 
SwiftNet can achieve 63.28\% top-1 accuracy on ImageNet-1K Classification Task with only 2 million parameters and 53 million MACs. 

We summarize our main contributions as follows:
\vspace{-2mm}
\begin{itemize}[noitemsep,leftmargin=*]
	\item We proposed \searchmethod, a new search algorithm for neural architecture based on meta-graph that can explore much more flexible search space and accumulate learned knowledge to enable high accuracy density search and good adaptability.
    \item We proposed a new structure-level pruning method, which can preserve high accuracy while transforming to small models, a significant improvement from the existing channel scaling methods. 	
    \item  By using the proposed search and pruning methods, the discovered set of models SwiftNet can outperform state-of-the-art NAS works in accuracy density, top-1 accuracy, latency, and MACs for small models.
    \item \searchmethod supports multi-objective neural architecture search and facilitates adding and changing objectives through changing the update rules in meta-graph. 
\end{itemize}

\begin{figure*}[!h]
\vspace{-0.3cm}
\setlength{\abovedisplayskip}{-3pt}
\setlength{\belowdisplayskip}{-3pt}
\includegraphics[width=\textwidth]{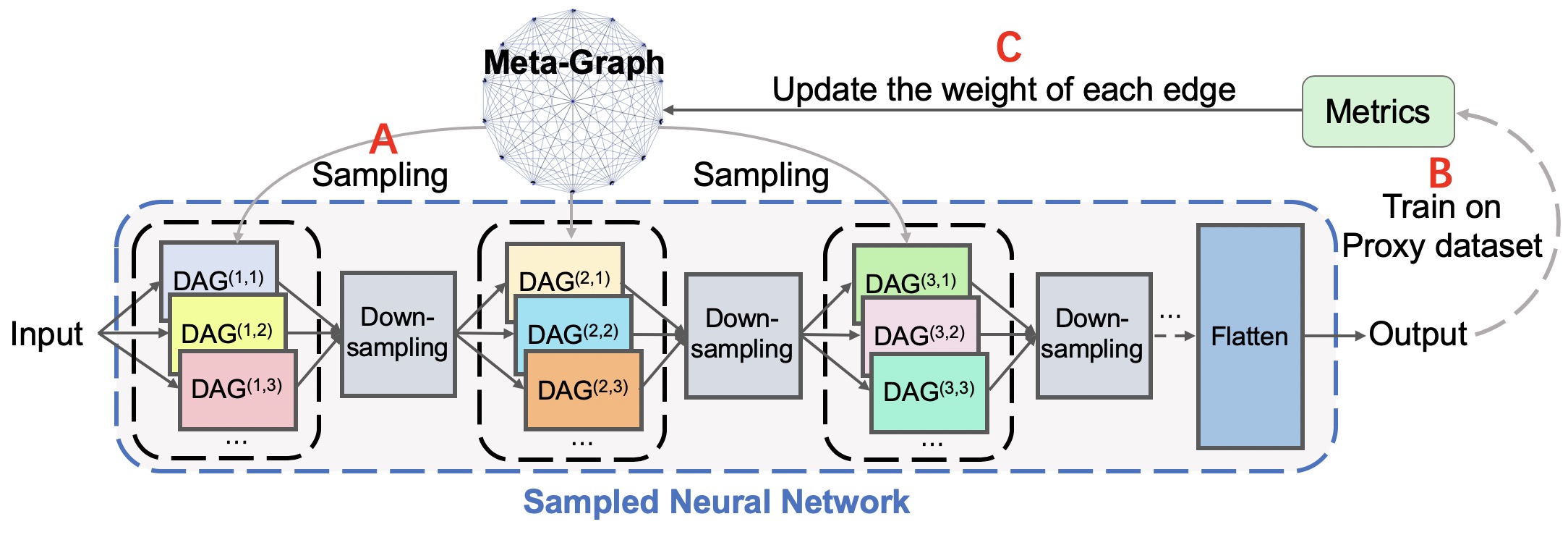}
\caption{Overview diagram of the search process. To form a sampled DNN, we subsample multiple DAGs from the complete DAG. After several training epochs with proxy training set, we use the search metrics (such as latency and accuracy) to update the complete DAG.}
\vspace{-1.0em}
\label{fig:workflow}
\end{figure*}

\section{RELATED WORK}


\paragraph{Deep Neural Networks for Mobile Devices:} 
Continuously increasing demand for the use of deep neural network in mobile device promotes the research for low cost and high-efficient models. In the past several years, mobile neural networks are well hand-crafted and achieved competitive performance. Squeezenet~\cite{iandola2016squeezenet} is a compact deep neural network designed by delicately selecting channel depth and the convolutional filters. For example, 3x3 convolution is replaced by 1x1 convolution.  Later on, Mobilenet-V1~\cite{howard2017mobilenets} is proposed and its separable depthwise convolution brought a big impact on efficient neural network design. Mobilenet-V2~\cite{sandler2018mobilenetv2} is a design with residual and bottleneck layers. These well tuned hand-crafted models have achieved breakthrough results on mobile vision tasks. These previous works provide us some good intuitions for designing good mobile models. To push the boundary further,
we need to efficiently explore unseen network design with novel operations. However, such dynamics makes it challenging and inefficient for manual design. 

\paragraph{Neural Architecture Search for Mobile Devices:}
After the idea of NAS~\cite{zoph2018learning, zoph2016neural} came out, we are able to find new neural architectures by automatically explore the combination of operations, activation functions, hyperparameters, etc~\cite{liu2018darts, enas, xie2018snas, tan2018mnasnet, yang2018netadapt}. While NAS achieved 74\% top-1 accuracy on ImageNet, it has 5.3 million parameters and 564 million MACs making it inapplicable to mobile devices. Recent works demonstrated to search mobile models by using FLOPs as regularization, considering inference latency into the search process and predicting the accuracy of candidate networks \cite{gordon2018morphnet, cai2018proxylessnas, wu2018fbnet, dai2018chamnet, tan2018mnasnet, he2018amc}. 

While these techniques help to find neural architectures with lower inference latency and computational cost, there are two main things that have not been well explored. First, to narrow down the wide search space, previous works used cells and blocks of layers to build neural architectures. However, this course-grained searching space induces a large number of MACs and parameters. 
Second, to downsize the found architecture to smaller a model, channel depth multiplier is generally adopted. By reducing the number of output channels, channel depth multiplier can dramatically decrease MACs. But it also comes with an inevitable and significant accuracy drop. For example, FBNet was proposed to find a mobile model with low latency and high accuracy. By using latency as an optimization factor combining with gradient-based search method, FBNet achieved 74.1\% top-1 accuracy on ImageNet, 296M FLOPs. However, to adapt to a smaller model (13.7 million MACs), FBNet's accuracy downgraded to 50.2\%~\cite{wu2018fbnet}.  Therefore, how to fast adapt found architecture to a smaller model for strict hardware constraints efficiently without much accuracy loss become vitally important. 

Recent progress in graph neural network~\cite{scarselli2009graph, li2015gated, kipf2016semi} demonstrates the potential of searching neural architecture with graph-based method~\cite{zhang2018graph}. For example, GHN~\cite{zhang2018graph} achieved 73\% top-1 accuracy on ImageNet in 0.84 GPU days. However, the neural architecture of GHN is randomly sampled and the searching process is unaware of structure redundancy. 
It causes the found model to have large MACs and low accuracy density as shown in Figure~\ref{fig:MAC_density} and~\ref{fig:acc_density}.

\section{METHODOLOGY}

We propose GRAM (GRAph propagation as Meta-knowledge), a DAG-based methodology for neural architecture search. 
Our main goal is to find the optimal neural architecture based on two search metrics: accuracy and latency~\footnote{We use accuracy and latency as example search metrics in this paper, but our method can be extended to support other search criteria.}:
\begin{equation}
    \mathcal{A}^{*} = \arg\max\limits_{\mathcal{A}} \mathcal{M}(\mathcal{A}),
    \label{con:2}
\end{equation}
where $\mathcal{A}$ represents a neural network structure and
\begin{equation}
    \mathcal{M} = Accuracy(w^{*}(\mathcal{A}), \mathcal{A}, \mathcal{D}) - \gamma \cdot Latency(\mathcal{A}).
\end{equation}
Here $Accuracy()$ denotes the accuracy of current architecture given optimal weight distribution evaluated by validation dataset. $Latency()$ is measured by the inference time of given network structure.

$\mathcal{D}$ represents the validation data set, $\gamma$ is an adjustable penalty term, and  
\begin{equation}
    {w^{*}(\mathcal{A})}=\arg\min\limits_{w}\mathcal{L}_{train}({w, \mathcal{A}}),
    \label{con:3}
\end{equation}
where $w$ represents trainable parameters (weights and biases) and $\mathcal{L}$ is cross-entropy loss without regularization.





\subsection{Graphical Representation of Architecture}





We represent the structure of the neural network using Directed Acyclic Graph (DAG): $G=(\mathcal{V,E})$. Each node $v \in V$ has an operation $o_v$, such as 1x1 convolution, parameterized by $w_v$. Each node produces an output tensor $x_v$. Edge $e_{u\rightarrow v} = (u, v, w_{uv})\in \mathcal{E}$ represents the flow of the tensor from node $u$ to node $v$, and the probability of the connection is given by $w_{uv}$. $x_v$ is computed by performing the corresponding operation on each input and then sum the output of operation as follows:

\begin{equation}
    x_v = \sum_{e_{u\rightarrow v} \in \mathcal{E}}o_v(x_u;w_v)
\end{equation}

\noindent {\bf Meta-Graph.} 
Every sampled neural network can be expressed as $K$ DAGs and each DAG is a sub-graph of a complete DAG. Therefore, we construct Meta-Graph as $K$ independent complete DAGs, which contains all the update information of sampled neural networks.



\subsection{Architecture Search using GRAM}

GRAM is an operation-wise architecture search method that can accumulate the knowledge learned in each update into meta-graph. Compared with existing NAS works, GRAM supports much wider search space with significantly fewer constraints. 

\noindent {\bf Search Space.}
Existing NAS works perform architecture search based on cell or block~\cite{wu2018fbnet} units. These units reduce the search complexity at the cost of flexibility. In this work, we use node as the basic search unit, where each node is one of the operations in the computational DAG. Node-based search is more flexible than cell-based or block-based search because it allows us to explore new cell structures rather than use the predefined modules.
For Convolutional Neural Networks (CNNs), the convolution is followed by batch normalization and ELU activation function~\cite{clevert2015fast}. Each node can choose the operation from below:
\begin{itemize}[noitemsep,leftmargin=*]
    \item 1 $\times$ 1 Convolution with 32, 64, 128 filters
    \item 3 $\times$ 3 Convolution with 32, 64 filters
\end{itemize}
In the interest of space, we use CNNs as an example throughout the paper to explain our approach, but it can be easily extended to other network structures by modifying the search space according to the requirements. For example, an extension to RNNs can be done by substituting the CNN operations in the meta-graph with RNN operations. 

For CNNs, down-sampling is an essential operation to reduce the sensitivity to output shifts and distortions \cite{lecun1998gradient}. 
Thus we first place several down-sampling operations in the neural network structure and use them as a coarse-grained hierarchy.
Then we fill multiple complete DAGs in parallel between 2 adjacent down-sampling operations to form a flexible search space. Multiple DAGs learn their structural representation independently to generate new architectures. Note that single DAG is feasible, but multiple DAGs can avoid the bias introduced in the random DAG institutionalization. 
The structure is visualized in Figure \ref{fig:workflow}. 
Each DAG represents a large space that contains many possibilities of the forwarding path. 
Each node can choose from 1 $\times$ 1 and 3 $\times$ 3 convolution with different channel numbers. The topology of complete DAGs provides all possible connections.
Therefore, our search space is much wider than other cell-based and block-based approaches. For example, with $n=30$ nodes in each complete DAG, $m=3$ DAGs in parallel between two down-sampling modules, and $h=3$ down-sampling modules in total, the search space contains $10^{84}$ possible architectures. In comparison, FBNet with block design only provides around $10^{21}$ possible architectures.


\noindent {\bf Search Algorithm.} Figure~\ref{fig:workflow} gives an overview of GRAM's search process. 
In phase A, we sample each DAG to obtain a sampled neural network. In phase B, we train the sampled neural network on proxy data set and get the performance metrics. Then in phase C, we update the connection weights in each DAG to meta-graph according to the metrics.
Since each DAG is sampled and updated individually, they are independent. Thus the optimal meta-graph distribution can be described as:
\begin{equation}
    P^{*}(\mathbf{E_w})=\prod_{k=1}^{K}p^{*}(\mathbf{E_{w}^{k}}),
    \label{tar:distr}
\end{equation}
where $E_{w}^k \in \mathbb{R}^{\frac{n(n-1)}{2}}$ is the connection weights of the k-th DAG, and $P(\mathbf{E_w})$ is the distribution function of the meta-graph. 

Each meta-graph is a sample of the optimal distribution. The optimal neural network structure $\mathcal{A}^{*}$ satisfying Eq.~\ref{con:2} is the mean of the optimal meta-graph distribution:
\begin{equation}
    \mathcal{A}^{*} = \mathbb{E}[f(e_w)]=\int f(e_w)p(e_w)de_w,
\end{equation}
where $f(e_w)$ is the probability density function of $\mathbf{E_w}$. Therefore, our meta-graph updating process is to approximate the mean of the distribution, which is in accordance with the goal of optimizing accuracy and latency.

Since there is no closed-form solution to compute the mean of the optimal meta-graph distribution, we adopt Gibbs Sampling method to approximate the optimal meta-graph by updating the connection weights in meta-graph. 
More specifically, at the beginning of each training round, we sample each complete DAG and construct the sampled neural network with the sampled DAGs and downsampling modules.
Let $E_w$ represent a sample in the optimal meta-graph, and the update process can be described as:
\begin{equation} \label{updateprocess}
\begin{split}
e_{w (i, j)}^{k, t + 1} \sim p(e_{w (i, j)}^{k} \given e_{w (0, 1)}^{k, t + 1}, e_{w (0, 2)}^{k, t + 1}, ...,  \\
e_{w (i - 1, j)}^{k, t + 1}, e_{w (i + 1, j)}^{k, t}, e_{w (i + 2, j)}^{k, t}, ...)\\
\forall{k\in [0, K)}, \forall{t\in [0, T)}, \forall{i,j} \in \mathcal{V}, i<j,
\end{split}
\end{equation}
where $K$ is the number of the DAGs, $T$ is the iteration times. The update process stops when meta-graph converges to the mean of the optimal meta-graph distribution and the convergence is guaranteed by Gibbs Sampling.
Adopting Gibbs Sampling instead of performing a search on the complete DAG also greatly reduces the computational overhead. 

\renewcommand{\algorithmicrequire}{ \textbf{Input:}}
\renewcommand{\algorithmicensure}{ \textbf{Output:}}
\begin{algorithm}[t]
\caption{Update Connection Weights in Meta-graph.}
\label{alg:completeDAG}
$\left. \textbf{Input:} \right.$ \newline
$\textbf{G:}$ Initialized meta-graph with uniform connection weights \newline
$\left. \textbf{Hyperparameter:} \right.$ \newline
$K$: total number of complete DAGs \newline
$n$: number of nodes for each DAG \newline
$m$: number of DAGs per hierarchy \newline
$h$: number of hierarchies in the architecture \newline
$\alpha$: convergence learning rate of meta-graph  \newline
$\beta$: moving expected performance on specific tasks, which is updated according to history performance collected through meta-graph training  \newline 
$\left. \textbf{Intermediate variables:} \right.$ \newline
$\tau$: latency of sampled model  \newline  
$\eta$: accuracy on the validation set of sampled model  \newline
$\eta'$: accuracy penalized by latency on the validation set of sampled model \newline
$Z_t$: Connection weight normalization term \newline 
\textbf{begin}
\begin{algorithmic}
\FOR{$t \gets 1$ to $T$}
\STATE Sub-sample Meta-graph and train on proxy dataset \newline \
$G_{sub} = subsample(\textbf{G})$ \newline 
    ${w^{*}} = \left. \arg\min\limits_{w}\mathcal{L}_{train}({w,G_{sub}, \mathcal{D}}) \right. $
\STATE Evaluate the model and calculate penalized accuracy 

    $\left. \eta , \tau = Eval(w^{*}, G_{sub}, \mathcal{D}) \right.$ 
    
    $\eta' = \left. \eta - \gamma \tau \right. $

\STATE Update connection weights for meta-graph:

\FOR {$k \gets 0$ to $K-1$}
\STATE Update connection weights for $DAG_k$ \newline \
    $e_{w(i,j)}^{k,t}=\left.\begin{cases}
      \frac{e_{w(i,j)}^{k,t-1}}{Z_t}
             \exp[\alpha(\eta'-\beta)] & \text{If $e^{k}_{i\rightarrow j}\in \mathcal{E}_{k}$}\  \\
      \frac{e_{w(i,j)}^{k,t-1}}{Z_t} & \text{Otherwise}
    \end{cases} \right. 
    $

\ENDFOR
\ENDFOR
 \end{algorithmic}
\textbf{end}
\end{algorithm}


\noindent {\bf Weights Update.}
After obtaining the sampled neural network structure, we train the network on the proxy data set and estimate the performance of the sampled network. In this work, we use accuracy and latency as example metrics for weight updating.
The weights are updated according to the difference between the estimated performance and expected performance.
Note that only the weights of edges chosen in the sampled neural network are updated. We apply exponential function to boost the updating rate, and a scale factor $\alpha$ to adjust the rate of the update. The updated weights are normalized for a valid probability representation. The pseudo code of the complete graph updating is given in Algorithm~\ref{alg:completeDAG}. 

\subsection{Structure-Level Pruning for Compact Model}
GRAM is highly adaptable to different sizes of models. To support resource-constraint edge computing applications, we propose a structure level pruning method.
Existing works such as MobileNets~\cite{howard2017mobilenets, sandler2018mobilenetv2} uses width multiplier to thinner the models and down-scale the channel depth to remove the redundant channels. 
Such approaches could not identify redundant operations and thus less efficient than the proposed structure-level pruning approach. 
To generate compact architectures from the meta-graph, we sample neural architectures by setting a pruning level for the connection weights. More specifically, the connection weights that are below the given level are pruned while the connections weights larger than the level are kept. 
In addition, we constrain the maximum filter depth in the search space, which shares the similar idea as channel down-scaling to efficiently reduce model size without accuracy loss.
We develop a profiling-based method to determine the pruning level based on the resource constraint, which is detailed in Section \ref{sec:performance_ana}.

\paragraph{Channel Up-scaling.}
Small models are prone to underfitting for challenging tasks because they have limited representation capacity.
For example, if we transfer the trained meta-graph on CIFAR-10 proxy dataset to the more challenging task ImageNet, the model capacity may be not sufficient and results in accuracy loss.
To solve this issue, we introduce the channel up-scaling technique. Channel up-scaling applies a scaling factor larger than 1 for each node's output channel. The channel up-scaling is node-wise in our approach, so the scaling has better flexibility.

\section{EXPERIMENTS}
\subsection{Experiment Setup}\label{sec:setup}


\textbf{Meta-graph.}
We configure meta-graph with 3 coarse-grained hierarchies. Each hierarchy contains 3 complete DAGs in parallel with 30 nodes for each DAG. Pooling size and stride for each down sampling layer is set to 2. We use batch normalization layer after each convolutional layer except the last one. We use ELU~\cite{clevert2015fast} activation after every batch normalization layer to speed up the training. 
$\alpha$ is set to 0.9. $\beta$ is the moving average of the history performance with initial value 0.4.
$\gamma$ is set to 1 to penalize latency in the unit of ms.
We trained the meta-graph for 1000 iterations.

\textbf{Proxy Dataset.}
Training the meta-graph on the entire dataset is too expensive, so we adopt the approach used in previous works \cite{tan2018mnasnet, zoph2018learning,real2018regularized, wu2018fbnet} to train the meta-graph on a small proxy dataset.
Here we randomly select 500 samples from the entire dataset as our proxy dataset.




\textbf{Full Model Training on CIFAR-10.}
We train the models generated from meta-graph with the full dataset for evaluating the performance, e.g., for CIFAR-10~\cite{krizhevsky2009learning}, there are 50,000 images in the training dataset and 10,000 images in the validation dataset. 
We use the momentum optimizer \cite{sutskever2013importance} with initial learning rate 0.01, and decay the learning rate by 0.1 at 50\% and 75\% of the total training epochs respectively. To prevent overfitting, we use 0.0005 L2 weight decay during training.

\textbf{Model Adaptation on ImageNet.}
After training the meta-graph on CIFAR-10, we adapt it to ImageNet-1K~\cite{deng2009imagenet} by adding sequential convolutional layers before the generated model and adjusting the final average pooling layer to match the dimension of the outputs. 
We adopt the hyper-parameter settings in Inception-v3 \cite{szegedy2016rethinking}, and use the RMSprop \cite{tieleman2012divide} optimizer with initial learning rate 0.045 and decay the learning rate by 0.94 for every two epochs. Since the generated model is not prone to overfitting, we use 0.00001 L2 weight decay during training to speedup the process. 



\begin{figure}[t]
\begin{center}
   \includegraphics[width=0.8\linewidth]{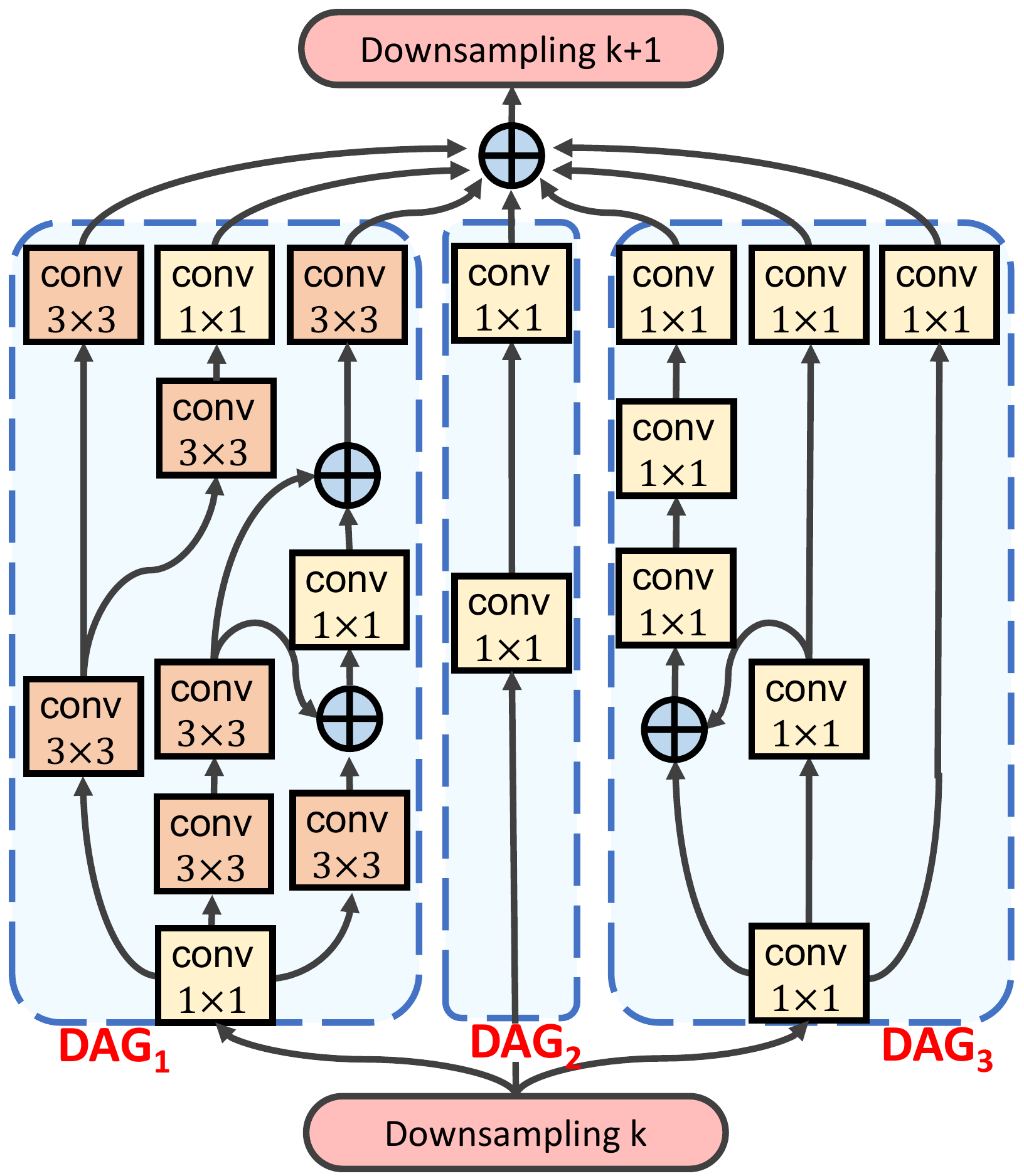}
\end{center}
\vspace{-1.2em}
   \caption{A representative architecture discovered by GRAM.}
   \vspace{-2mm}
\label{fig:onecell}
\end{figure}

\subsection{Performance Analysis}
\label{sec:performance_ana}

\paragraph{SwiftNet.}
Figure~\ref{fig:onecell} demonstrates a representative architecture (composed of DAGs as building components) discovered by GRAM and we name it SwiftNet. 
With different initialization, each DAG develops its independent structure after graph propagation through meta-graph training and structure-level pruning.
The structure knowledge learned by each DAG is highly representative: the DAG on the left learns a combination and concatenation of different-sized filters to extract spatial information from the input feature map; the DAG in the middle identities mapping to transfer the current input feature map to the next hierarchy; and the DAG on the right explores a set of $1 \times 1$ convolutional operators. Intuitively, these $1 \times 1$ operators serve as the approximation of $3 \times 3$ convolutional filters to best utilize the information from the input feature map.

As shown in Table~\ref{tab:CIFAR-Prune}, compared with state-of-the-art methods, SwiftNet has much higher accuracy density while fewer parameters, i.e., 85.92\% with only 250k parameters.
Unlike FBNet~\cite{wu2018fbnet} and other cell-based or block-based approaches, SwiftNet is not restricted to stacking sequential layers with mutable channel depth and negligible connections and thus has much larger search space and more flexible architecture choice.
SwiftNet contains the characteristic of Inception and Densely-connected network \cite{szegedy2016rethinking, huang2017densely}.
The Inception-like structure is enabled by paralleling different DAGs to extract and map the input feature maps, and DenseNet-like structure is enabled by preserving the node connections in each DAG. 
SwiftNet is less memory and computationally intensive than Inception and DenseNet as the structure-level pruning heavily reduces redundant connections in the meta-graph while preserves critical filter stacks and concatenations.

\begin{table}[t]
\centering
\caption{Performance and accuracy density of different versions of SwiftNet trained on CIFAR-10 in the format of SwiftNet-0.5, where 0.5 represents the level of structure-level pruning.}
\scalebox{0.9}{
\begin{tabular}{ccccc}
\hline

Model & Accuracy & \#MACs & Density\\
      & (\%)     &            & (\%/M-MACs)\\
\hline
SwiftNet-0.65 & 85.92 & 3M & 28.64\\
SwiftNet-0.5 & 88.74 & 5M & 17.75\\
SwiftNet-0.4 & 89.09 & 12M & 7.42\\
SwiftNet-0.3 & 90.17 & 36M & 2.50\\
\hline
   \vspace{-5mm}
\end{tabular}}
\label{tab:CIFAR-Prune}
\end{table}


\begin{figure}[b]
\begin{center}
   \includegraphics[width=1.1\linewidth]{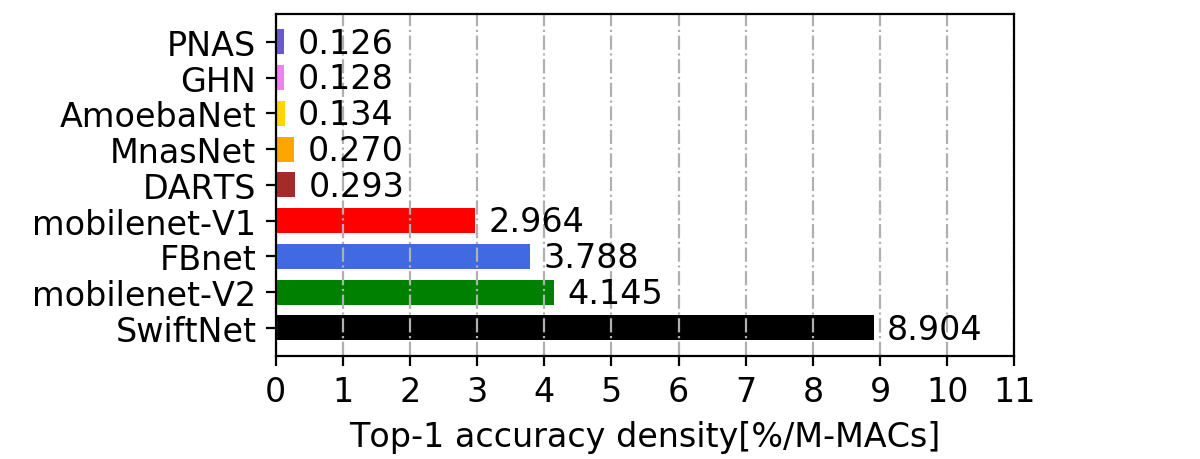}
   \end{center}
   \vspace{-1em}
   \caption{ImageNet-1K top-1 accuracy density comparison between SwiftNet and state-of-the-art NAS approaches. Accuracy density shows how well a model use computational resource. SwiftNet achieves 2.15$\times$ accuracy density compared with the MobileNets.}
\label{fig:acc_density}
\end{figure}

\begin{table*}[t]
\centering
\caption{Compact model comparison between SwiftNet and baselines on ImageNet-1K. The 0.35 in MobileNet-V2-0.35 and FBNet-96-0.35-2 both stand for 0.35 width multiplier. In comparison, SwiftNet-0.4 does not applied width multiplier. By using structure pruning with a level of 0.4, SwiftNet achieves a better accuracy and latency trade-off. *Our searching is done by 1 NVIDIA GEFORCE GTX 1080. $\dagger$ Since our work is focusing on compact model design, we compare SwiftNet with baseline models of similar scale. 
}
\begin{tabular}{ccccccc}
\hline
Model & Method & Search   & Search Cost & Latency(ms) &\#MACs &   Top-1      \\ 
              &  & Space    & (GPU hours) &&  & Accuracy(\%)\\ \hline
MobileNet-V2-0.35$\dagger$& Manual &--  & -- & 4.5 & 11M & 45.5\\
FBNet-96-0.35-2$\dagger$ & Gradient & layer-wise & 216 & 4.26 & 13.7M & 51.9\\
\textbf{SwiftNet-0.4}$\dagger$ & \textbf{\searchmethod} & \textbf{node-wise} &   \textbf{8.3}* & \textbf{4.24} & \textbf{10.4M} & \textbf{54.4}\\
\hline

\end{tabular}

\label{tab:comparison-performance}
\end{table*}
   \vspace{-7mm}
\paragraph{Performance Comparison with State-of-the-art NAS Methods.}

Figure~\ref{fig:acc_density} illustrates that SwiftNet outperforms other architectures in terms of accuracy density, which indicates that SwiftNet can utilize the parameters and MACs more efficiently.
Table~\ref{tab:comparison-performance} shows the comparison between SwiftNet and state-of-the-art NAS approaches in terms of search cost, latency, and top-1 accuracy. In the interest of space and also the fact that some NAS approaches mainly focus on relatively large models (i.e., they do not provide the results of comparable small models), the results of NAS approaches with significantly lower accuracy density are not shown in the table. 
The results show that SwiftNet outperforms both MobileNet-V2 and FBNet in all metrics. 
Specifically, the search speed of GRAM is about $26 \times$ faster than FBNet even though the search space of GRAM is more than $10^{63}$ times larger, which suggests the high search efficiency of GRAM. The high search efficiency comes from the sampling process and weight update scheme as connections that lead to bad performance on proxy dataset are unlikely to be sampled in the next iteration.
In terms of accuracy, our node-wise search space allows GRAM to explore models with greater structure diversity than MobileNet-V2 and FBNet. In a vision perspective, by combining both simpler and more complex structures from independent DAGs, our meta-graph is able to approximate hypercomplex receptive fields \cite{glezer1973investigation} in favour of accuracy.
Detailed discussion on the latency advantage of SwiftNet is present in Section~\ref{sec:latency_analysis}.

\begin{figure}[b]
\begin{center}
   \includegraphics[width=\linewidth]{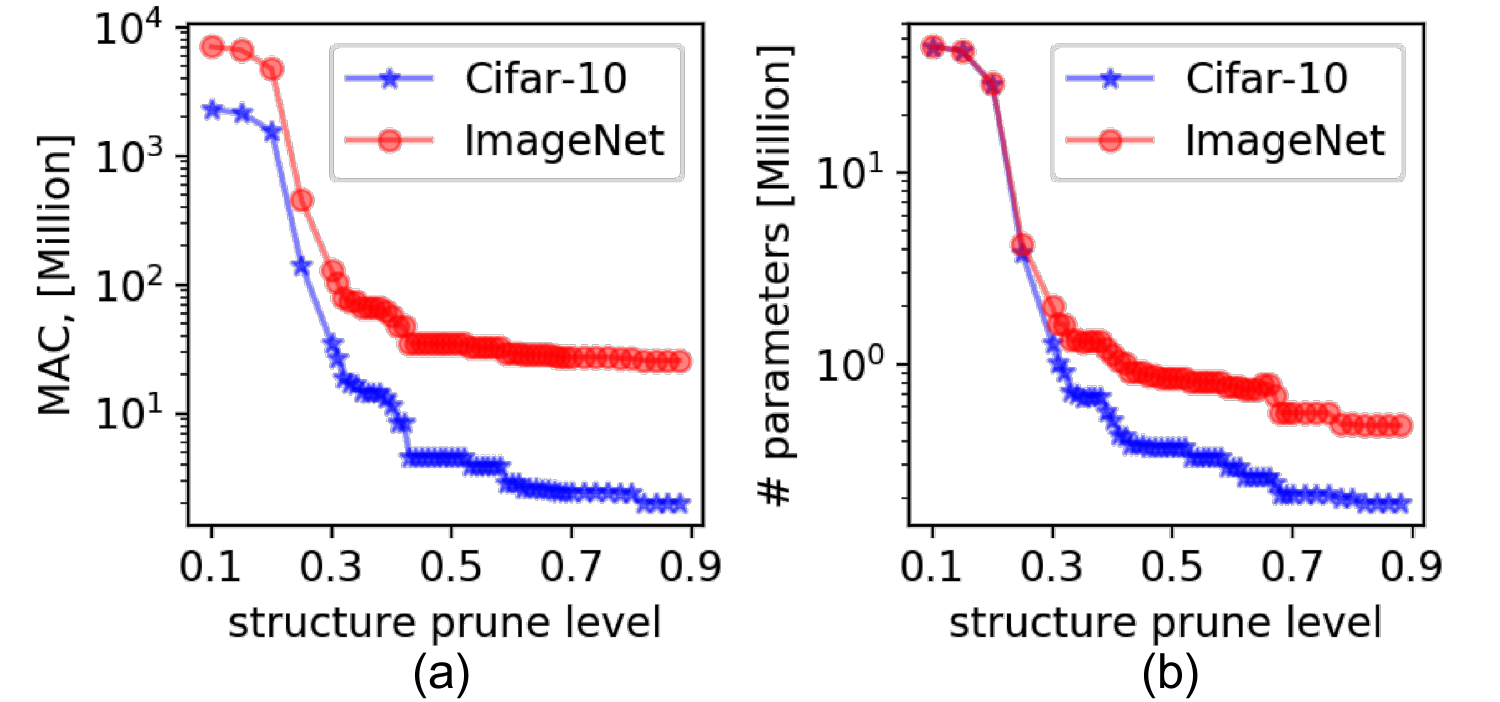}
\end{center}
   \vspace{-1.3em}
   \caption{Model scale vs Structure pruning level.  
}
\label{fig:pruning}
\end{figure}

\paragraph{Structure-Level Pruning.}
The proposed structure-level pruning method is used to extract and refine the final model according to the resource constraints or other objectives. 
To determine the pruning level based on the resource constraint, we conduct profiling to find out the relationship between model scale (or other resource constraint metrics) and structure pruning level. We convert our sampled architectures to TensorFlow models and do the profiling to calculate MACs and number of parameters.
The profiling is lightweight as it can be finished in less than 1 hour on a Xeon E5 CPU.
Figure~\ref{fig:pruning} shows the profiling results. 
The model scale drops down drastically in the range of [0.2, 0.6], which suggests pruning is more efficient in this range.

\paragraph{Channel Up-scaling.}

We evaluate the same generated architecture with channel up-scaling factor 1.25 and 1.5 on ImageNet classification task. To ensure a fair comparison between different scaling factors, we use fixed data-preprocessing pipelines and hyper-parameter settings in training process. 
Table~\ref{tab:performance-SwiftNet} demonstrates the accuracy, MACs, and Model Size under different scaling settings on ImageNet. 
As expected, the best model has the lower pruning level and higher up-scaling factor.
The results suggest Channel Up-Scaling can provide up to 5\% top-1 accuracy improvement at the cost of about 1 Million additional parameters.
This verifies that Channel Up-Scaling is useful when transfer existing model for new tasks, especially when the new task is more challenging than existing model.  

\begin{table*}[t]
\centering
\caption{Performance, MACs, and Model Size of different versions of SwiftNet trained on ImageNet-1K in the format of SwiftNet-96-0.5-1.00, where 96 represents the image input size, 0.5 represents the structure prune level, and 1.00 represents the channel up-scaling factor. Channel Up-scaling provides up to 5\% performance gain at the cost of up to 2$\times$ model size.}
\scalebox{0.95}{
\begin{tabular}{ccccc}
\hline
Model & Top-1 Accuracy(\%) & Top-5 Accuracy(\%) & \#MACs & \#Parameter \\ \hline
SwiftNet-96-0.5-1.00 & 48.24 & 73.56 & \textbf{10M} & \textbf{0.87M}\\
SwiftNet-96-0.4-1.00 & 54.44 & 77.12 & 14M & 1.15M\\
SwiftNet-128-0.5-1.00 & 54.69 & 77.27 & 18M & \textbf{0.87M}\\
SwiftNet-128-0.4-1.00 & 59.13 & 81.32 & 26M & 1.15M\\
\hline
SwiftNet-96-0.5-1.25 & 51.03 & 74.33 & 15M & 1.20M\\
SwiftNet-96-0.4-1.25 & 56.21 & 78.83 & 21M & 1.58M\\
SwiftNet-128-0.5-1.25 & 57.91 & 80.51 & 26M & 1.20M\\
SwiftNet-128-0.4-1.25 & 61.99 & 84.26 & 38M & 1.58M\\
\hline
SwiftNet-96-0.5-1.50 & 52.76 & 76.01 & 20M & 1.56M\\
SwiftNet-96-0.4-1.50 & 57.46 & 80.03 & 30M & 2.07M\\
SwiftNet-128-0.5-1.50 & 59.45 & 81.62 & 36M & 1.56M\\
SwiftNet-128-0.4-1.50 & \textbf{63.28} & \textbf{85.22} & 53M & 2.07M\\
\hline
   \vspace{-7mm}
\end{tabular}}
\label{tab:performance-SwiftNet}
\end{table*}


\begin{figure}[b]
\begin{center}
   \includegraphics[width=0.85\linewidth]{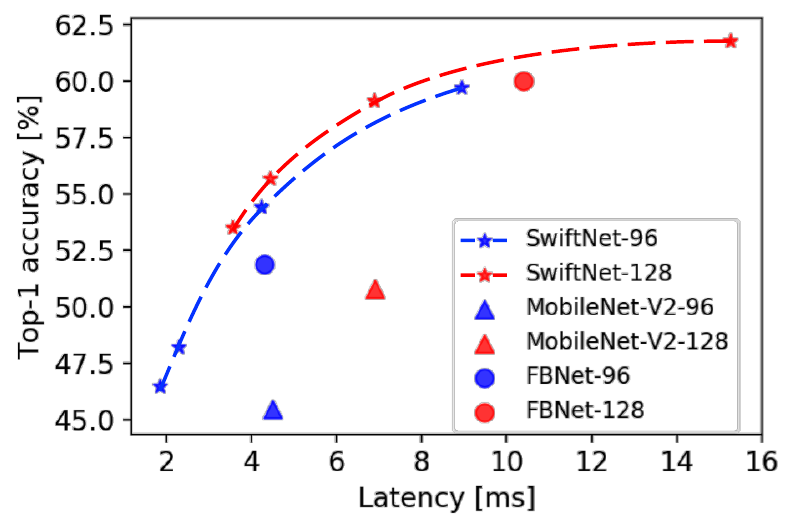}
\end{center}
   \vspace{-1.3em}
   \caption{Trade-off between latency and accuracy on ImageNet-1K Classification Task. We achieve significant latency improvement than MobileNet-V-2 and FBNet under the same top-1 accuracy, i.e., up to 2.42 and 1.47 speedup respectively. Since FBNet did not report latency on Google Pixel 1, we map FBNet's latency to Google Pixel 1 according to MACs by referring to~\cite{wu2018fbnet}.} 
\label{fig:runtime}
\end{figure}

\subsection{Latency Analysis}\label{sec:latency_analysis}
To measure runtime latency, we prototype SwiftNet as a TensorFlow Lite model and deploy it as a mobile APP on the Google Pixel 1. We compare the accuracy-latency curve between SwiftNet and MobileNet-V2 \cite{sandler2018mobilenetv2} under input size of 96$\times$96 and 128$\times$128 in Figure~\ref{fig:runtime}. 
Under the same accuracy, the latency of SwiftNet is significantly better (i.e., up to 2.42$\times$ better than MobileNet and up to 1.47$\times$ better than FBNet) for input size of 96$\times$96 and is also better (i.e., up to 2.11$\times$ better than MobileNet and up to 1.12$\times$ better than FBNet) for input size of 128$\times$128.
In addition, our experimental results also show that the architecture of SwiftNet makes good use of the parallel scheme, i.e., SwiftNet gets about 2$\times$ speedup in runtime latency when running in multi-threading mode than single-thread.  
This is because the computation of the three DAGs can be performed concurrently, see Figure~\ref{fig:onecell}.

\begin{table}[b]
    \centering
    \caption{Adaptability of meta-graph to different tasks.}    
    \scalebox{0.9}{
    \begin{tabular}{c|c|c|c}
         \hline
         Task  & \#MACs & Accuracy & Accuracy Density \\
               &        &   (\%)   &  (\%/M-MACs)\\
         \hline
         MNIST           & 0.23M & 99.24 & 431.47 \\
         SVHN            & 1.15M & 94.95 & 82.56\\
         Fashion MNIST   & 0.86M & 93.56 & 108.79\\
         \hline
    \end{tabular}}

    \label{tab:adaptation}
\end{table}

\subsection{Structural Adaptability to Different Tasks}

One of the most important contributions of the proposed method is that the trained meta-graph can be easily adapted to new tasks and schemes structurally. 
Existing neural architecture search methods finds a definite architecture with very limited degree of freedom as the found architecture can only be changed by applying the width multiplier or changing layers around the searched architecture.
Although changing the filter depth is beneficial to adjusting the model to new tasks, the models usually cannot be structurally adapted to the new scheme. Therefore, they usually require multiple steps of updates (shots) to tune the architecture towards the new task. 

However, our GRAM methodology is capable of providing a range of flexible architectures by changing the level of structure-level pruning. Since structural information is preserved and accumulated during meta-graph training, we no longer need to do the fine-tune updates before adapting our meta-graph to new tasks. This means it is convenient for meta-graph to directly form a new deep neural network for new tasks without further training.

We select a wide range of image classification tasks and use the trained meta-graph (trained on CIFAR-10) to directly generate architectures for them. The datasets include MNIST \cite{lecun1998gradient}, Fashion MNIST \cite{xiao2017fashion}, and SVHN \cite{netzer2011reading}. 
Table~\ref{tab:adaptation} shows the performance of the generated architectures. The adapted models achieve good performance (i.e., 99.24\% accuracy on MNIST, 94.95\% accuracy on SVHN, and 93.56\% accuracy on Fashion MNIST) while preserving at least 80 accuracy density, thanks to the great flexibility in the trained meta-graph.

\subsection{Limitations of GRAM}
While GRAM is able to explore a large amount of potential neural architectures in a short amount of time, the convergence of meta-graph relies on Gibbs Sampling, which does not provide a strict guarantee of upper bound with respect to the number of iterations before convergence. A rigorous theoretical guarantee on meta-graph convergence analysis is deferred as our future work.
\section{CONCLUSION}
We present GRAM (GRAph propagation as Meta-knowledge) for searching highly representative neural architectures. 
By using meta-graph as meta-knowledge to preserve learned knowledge, GRAM can efficiently search a space that is $10^{63}$ larger than FBNet's search space in just $8.3$ hours.
Unlike previous works, instead of using channel depth scaling for the trade-off between resource and accuracy, we propose structure-level pruning to achieve a higher accuracy density. GRAM successfully discovered a set of models named SwiftNet, which outperforms MobileNet-V2 by $2.15 \times$ higher accuracy density, 2.42$\times$ faster with similar accuracy. SwiftNet also outperforms FBNet, a differentiable-based searching method, by $26 \times$ in search cost and $1.47 \times$ lower in inference latency while preserving a similar accuracy. 
SwiftNet can achieve 63.28\% top-1 accuracy on ImageNet-1K Classification Task with only 2 million parameters and 53 million  MACs. 
GRAM can be strengthened by the proposed channel up-scaling to avoid underfitting when adapt to more challenging tasks using trained meta-graph. 
GRAM can be quickly adapted to different tasks with no update shot, thanks to the flexibility of proposed meta-graph.



{\small
\bibliographystyle{ieeetr} 
\bibliography{egbib}

}

\end{document}